%% file: main.tex
\documentclass[conference]{IEEEtran}
\IEEEoverridecommandlockouts

\usepackage{cite}
\usepackage{amsmath,amssymb,amsfonts}
\usepackage{graphicx}
\usepackage{textcomp}
\usepackage{xcolor}
\usepackage{booktabs}
\usepackage{multirow}
\usepackage{array}
\usepackage{url}
\usepackage{siunitx}

\begin{document}

\title{DynLimit: A Simulation Benchmark for Dynamics-Critical Planning and Control}

\author{\IEEEauthorblockN{1\textsuperscript{st} Yang Xiao}
\IEEEauthorblockA{\textit{school of Mechanical Engineering} \\
\textit{Southeast University}\\
Nanjing, China \\
230258029@seu.edu.cn}
\and
\IEEEauthorblockN{2\textsuperscript{nd} Yongzhi Liu}
\IEEEauthorblockA{\textit{school of Mechanical Engineering} \\
\textit{Southeast University}\\
Nanjing, China \\
230268037@seu.edu.cn}
\and
\IEEEauthorblockN{3\textsuperscript{rd} Sunan Zhang}
\IEEEauthorblockA{\textit{school of Mechanical Engineering} \\
\textit{Southeast University}\\
Nanjing, China \\
sunzhang@seu.edu.cn}
\and
\IEEEauthorblockN{4\textsuperscript{th} Kang Wang}
\IEEEauthorblockA{\textit{school of Mechanical Engineering} \\
\textit{Southeast University}\\
Nanjing, China \\
wangkang\_seu@seu.edu.cn}
\and
\IEEEauthorblockN{5\textsuperscript{th} Guodong Yin}
\IEEEauthorblockA{\textit{school of Mechanical Engineering} \\
\textit{Southeast University}\\
Nanjing, China \\
ygd@seu.edu.cn}
\and
\IEEEauthorblockN{6\textsuperscript{th} Weichao Zhuang*}
\IEEEauthorblockA{\textit{school of Mechanical Engineering} \\
\textit{Southeast University}\\
Nanjing, China \\
wezhuang@seu.edu.cn}
}

\maketitle

\input{sections/00_abstract}
\input{sections/01_introduction}
\input{sections/02_related_work}
\input{sections/03_platform}
\input{sections/04_protocol}
\input{sections/05_experiments}
\input{sections/05_conclusion}

\bibliographystyle{IEEEtran}
\bibliography{refs}

\end{document}

%% file: sections/00_abstract.tex
\begin{abstract}
Experiments near vehicle handling limits must keep three questions apart: what a maneuver requests, what the closed-loop system executes, and whether the task is completed.
We present DynLimit, a simulation platform that organizes this process end to end.
Scenarios are composed from road, friction, obstacle, and reference objects; planners and controllers attach through a shared observation and control interface; a vehicle adapter runs either a license-free four-wheel model or an external dynamics solver; and synchronized records feed task evaluation, replay, and response inspection.
The platform carries a task-oriented evaluation method that pairs a prescribed reference demand with a sampled passing set and event-aligned diagnosis.
Across eight task families, three friction environments, and three reference-assisted baseline chains we executed 450 runs: 360 nominal, 18 locally refined, and 72 with prescribed initial offsets.
Method differences depend on the task family, equal aggregate pass counts arise from different passing cells, and local refinement narrows two sampled transitions without locating a continuous boundary.
Two case comparisons show why sideslip or tire-utilization peaks cannot replace task constraints.
The three views therefore stay distinct in every reported run: an outcome can be traced from the prescribed demand, through the executed response, to the task rule that decided it.
Code is available at \url{https://github.com/Yang2581/dynlimit}.
\end{abstract}

%% file: sections/01_introduction.tex
\section{Introduction}
Evaluation of planning and control near the vehicle handling limit rests on three kinds of evidence that are easily conflated.
First, the prescribed maneuver fixes what the road and the reference demand: a speed profile, a curvature, and the resulting accelerations.
Second, the closed loop yields the executed response: the vehicle states and tire forces produced, together with the commands applied to obtain them.
Third, the task rule decides the outcome: whether the vehicle met its goal without leaving the road, striking an obstacle, missing a gate, or exceeding its deadline.
Because the three are recorded on different objects, none follows from another, so a pass count alone neither explains a failure nor locates the condition at which behavior changes.
The two same-scenario pairs in Section~\ref{sec:case} show how far the views can diverge: in the first, the chain with the larger peak sideslip completes the task while the other violates a geometric constraint.

Obtaining this evidence repeatedly is a systems-engineering problem as much as a modelling one.
Each trial is assembled from road geometry, a friction field, obstacles, an initial state, reference information, and an algorithm.
The loop must run at declared rates, with controls and states recorded on a common timeline, and each outcome must be decided by the same task rules online and offline.
Simulators supply vehicle models and scenario formats, and planning benchmarks supply scenario sets and metrics~\cite{dosovitskiy2017carla,okelly2020f1tenth,remonda2024assetto,althoff2017commonroad,karnchanachari2024nuplan}.
The layer between them, an experiment process that keeps these pieces aligned across tasks and algorithms, is still reassembled for most studies and has only recently begun to be automated in parts~\cite{lambertini2026sii}.

This paper presents \emph{DynLimit}, a simulation platform for dynamics-critical planning and control experiments, together with the task-oriented evaluation method embedded in it.
DynLimit is the platform: it owns the scenario objects, the algorithm and vehicle interfaces, the execution schedule, and the records.
\emph{DynLimit-Bench} is the method embedded in that process: reference demand organizes the test set, goals and constraints decide outcomes, and event-aligned records support diagnosis.
The quantitative experiments use reference-assisted control, in which every chain receives the same task description and reference while a shared longitudinal controller supplies speed commands; the platform interface itself is not restricted to this setting.

The contributions are threefold.
\emph{Platform design for dynamics experiments:} one experiment interface connects scenario composition, algorithm attachment, vehicle adaptation, execution scheduling, and record inspection, and we describe the interfaces and trade-offs behind it, including how interactive and scripted operation share scenario, controller, and vehicle components.
\emph{A task-oriented evaluation method embedded in that process:} each record is reduced to a demand-indexed passing set and a first-failure diagnosis, with stated sampling and tie-breaking conventions.
\emph{Systematic experiments and case analyses:} eight task families, 360 nominal runs, and 90 supplementary runs show batch comparison, local refinement of sampled transitions, and inspection of failures in the recorded response.
All quantitative statements are conditional on the declared model, configurations, and finite samples.

%% file: sections/02_related_work.tex
\section{Related Work}
\subsection{Dynamics-Aware Simulation Platforms}
Platforms for closed-loop driving research differ chiefly in vehicle-model fidelity and in how much of the evaluation they take on.
CARLA provides an urban simulator with configurable sensors and controlled scenarios~\cite{dosovitskiy2017carla}.
F1TENTH combines one-tenth-scale hardware with virtual racing tasks~\cite{okelly2020f1tenth}, and AssettoCorsaGym exposes a commercial racing simulator to reinforcement learning and model predictive control~\cite{remonda2024assetto}.
R-CARLA couples a learning environment with interchangeable vehicle dynamics~\cite{brunner2025rcarla}, while multibody solvers such as CarSim supply detailed tire and suspension models for handling studies~\cite{carsimMathModels}.
These systems show that closed-loop simulation can be scripted across model fidelities.
How a result should be read still depends on that model: simplifications change closed-loop trajectory following at different margins to the acceleration limit when assessed against real driving data~\cite{iv2024fidelity}, so declaring the vehicle, actuator, and simulation settings is part of the measurement.

\subsection{Scenario Organization and Task Evaluation}
Work organized around scenarios differs instead in what it fixes: the problem statement and its metric, or the scene alone.
CommonRoad makes motion-planning experiments composable through vehicle models, cost functions, and scenarios with goals and constraints~\cite{althoff2017commonroad}.
Scenic describes distributions over scenes with declarative constraints~\cite{fremont2019scenic}, MetaDrive and SMARTS organize traffic interaction for learning and evaluation~\cite{li2022metadrive,zhou2021smarts}, and nuPlan pairs recorded driving with closed-loop planning assessment that accounts for surrounding traffic~\cite{karnchanachari2024nuplan}.

Either way, these works fix the problem and the scene before evaluating planners at scale.
The question we pursue is narrower and procedural: for a prescribed maneuver family under a declared vehicle and controller configuration, which tested conditions are completed, where do sampled outcomes change, and what response precedes each failure?
Answering it requires the scenario objects, the algorithm interface, the execution schedule, and the task rules to stay consistent from run to run.
DynLimit is organized around that requirement, and the method it carries reports finite sampled passing sets rather than a capability score.

%% file: sections/03_platform.tex
\section{The DynLimit Platform}
\label{sec:platform}

\begin{figure*}[t]
\centering
\includegraphics[width=\textwidth]{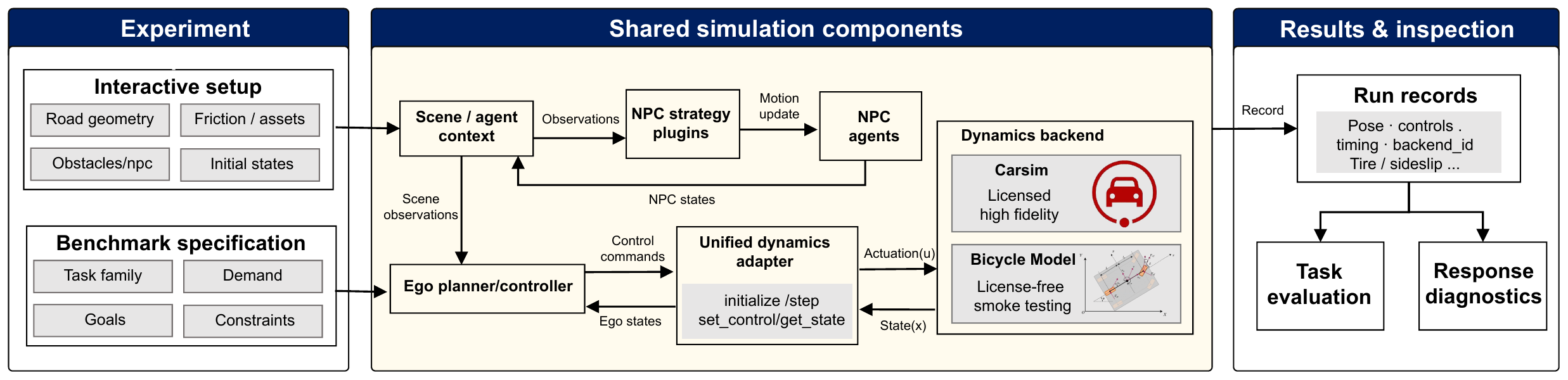}
\caption{DynLimit architecture.
Interactive setup (road, friction, assets, obstacles, initial states, agent policies) and benchmark specification (task family, demand, goals, constraints) define an experiment.
Shared simulation components (ego planner or controller, dynamics adapter, motion update) execute it on a selectable backend, and the resulting records feed task evaluation, response diagnostics, and offline inspection.
}
\label{fig:architecture}
\end{figure*}

DynLimit organizes a dynamics experiment as a configuration, a closed-loop execution, and a record (Fig.~\ref{fig:architecture}).
Both experiment entries, the interactive session and the programmatic batch, call the same scene, controller, and vehicle components.

\subsection{Architecture and Experiment Interfaces}
\label{sec:arch}

\begin{figure*}[t]
\centering
\includegraphics[width=\textwidth]{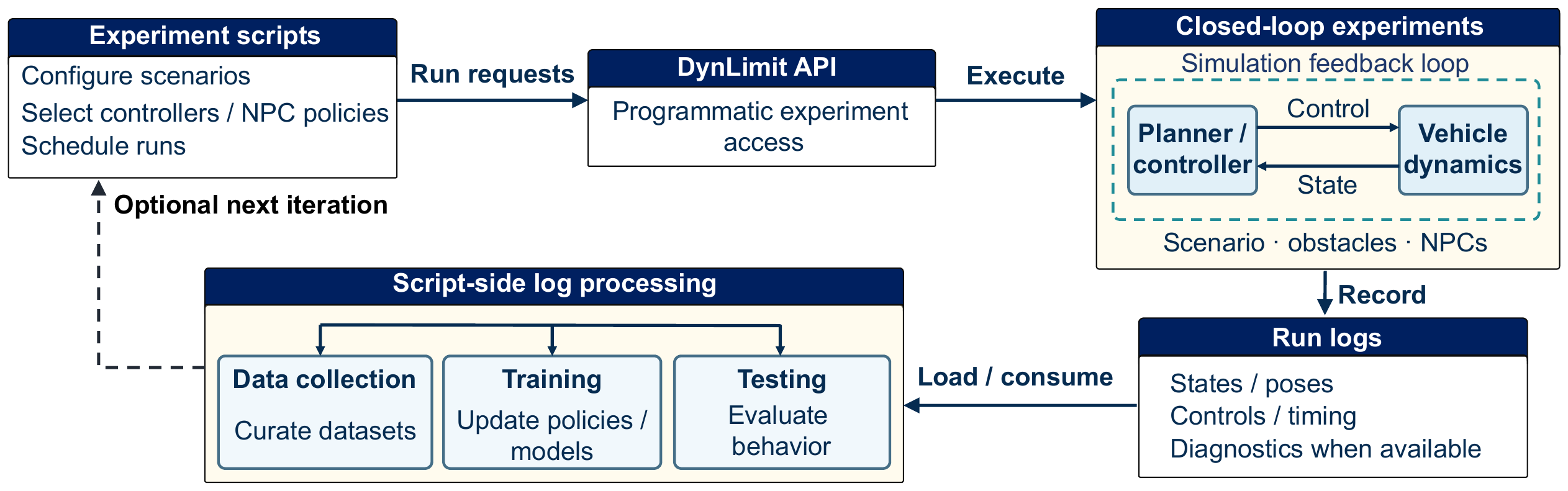}
\caption{Programmatic entry.
A specification generates a task matrix over families, levels, and chains; the scheduler executes it headlessly and writes one record per run; evaluation, diagnostics, and replay consume the records, and a revision returns to the matrix.
The excerpt at right shows the per-run labels used for the batches of Section~\ref{sec:experiments}.
No operator input enters between runs, so a repeated matrix reproduces the same records from the same declared entries.
}
\label{fig:loop}
\end{figure*}

An experiment is described by declared items: road geometry, friction field, initial vehicle state, reference information, algorithm identifier, vehicle configuration, and execution settings.
A task adds a completion goal, geometric constraints, and a deadline.
The platform treats all items as data; nothing about a family or level is hard-coded in the loop, which is what allows one execution path to serve the eight families of Section~\ref{sec:protocol}.
The two entries serve different purposes: interactive operation lets a scenario be examined while it is built and while it runs, whereas scripted operation executes a task matrix without supervision and without operator-dependent variation (Fig.~\ref{fig:loop}).
Both produce the same record type, so an interactive result can be re-evaluated by the same rules as a batch result.

The interface admits more than these experiments exercise.
This paper quantifies one backend, one entry, and three algorithm chains; the UI session, the HTTP interface, interactive agents, and the external solver integration are capabilities we describe as designed behavior, and Table~\ref{tab:evidence} marks which the reported runs support.

\subsection{Scenario Editing and Interactive Operation}
\label{sec:interactive}

Editing is object-based.
Road polygons, reference polylines, obstacles, and friction regions are separate entries with independent coordinates, so a lateral avoidance reference does not move the road boundary and a friction boundary may cross a lane at an arbitrary angle.
This is the platform-side counterpart of the task design in Section~\ref{sec:tasks}, where the same constructs become sweep variables and constraints.

\begin{figure*}[t]
\centering
\includegraphics[width=\textwidth]{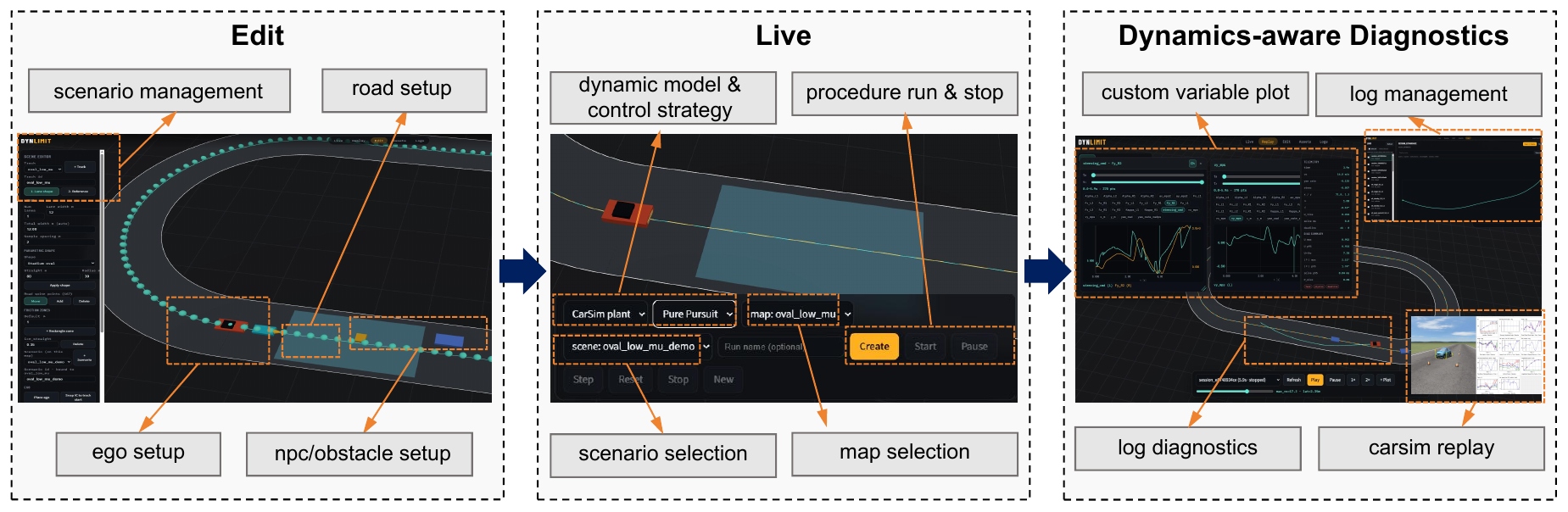}
\caption{Interactive interface panels on one task.
Setup defines ego, road, and obstacle objects and selects the map and scenario; run control starts and stops a closed-loop session with the selected dynamics model and control strategy; the diagnostics panel plots custom state variables and logged channels for replay.
The panels illustrate platform capability; the quantitative results use the scripted entry.
}
\label{fig:interactive}
\end{figure*}

Three modes cover the experiment cycle (Fig.~\ref{fig:interactive}).
\emph{Edit} fixes the task: geometry, friction, initial state, and the reference that defines the demand.
\emph{Live} executes a closed-loop session on the selected backend and streams ego states and controls, obstacle and agent poses, and controller diagnostics.
\emph{Replay} presents recorded trajectories and available dynamics channels on a shared timeline; visualization is decoupled from the update loop, so playback does not alter the record.
Interactive agents with pluggable policies extend the same representation to multi-vehicle behavior; no result below uses them.

\subsection{Algorithm and Vehicle Integration}
\label{sec:integration}

An algorithm receives a shared observation (pose, body-frame velocity, yaw rate, reference geometry, configured obstacles) and returns a road-wheel steering angle with normalized throttle and brake commands, held between updates.
Because the observation does not depend on the algorithm, chains can be exchanged within a fixed task without editing the scenario, which is the property the comparison in Section~\ref{sec:main-results} relies on.

The vehicle side is an adapter with initialization, reset, control application, stepping, state retrieval, capability reporting, and shutdown.
An external solver attaches through the same adapter, using a session configuration generated from working copies of its parameter and path files; the CarSim integration is documented but not exercised below.

Every reported run uses the same bicycle dynamic model, which carries longitudinal, lateral, and yaw body motion, four wheel-spin states, and a steering-actuator state.
Pure-slip tire curves are followed by an isotropic combined-force projection,
\begin{equation}
(F_x,F_y)=\frac{(F_x^0,F_y^0)}{\max\!\left(1,\sqrt{(F_x^0)^2+(F_y^0)^2}/(\mu F_z)\right)},
\end{equation}
and normal loads include quasi-static longitudinal and lateral transfer.
The nominal vehicle has 1500 kg mass, 2500 kg\,m$^2$ yaw inertia, axle distances of 1.2 m and 1.6 m, a 1.6 m track, and a 0.55 m center-of-mass height; steering has a 0.05 s time constant, a 0.8 rad/s plant rate limit, a 0.6 rad magnitude limit, and a 1 rad/s command rate guard.
The model is compact and uncalibrated on purpose: it provides an identical dynamic plant for every chain rather than reproducing a particular car.

Algorithms attach as plug-ins.
The three chains used here are Pure Pursuit, Stanley, and a lattice planner with longitudinal PID, all configured before the matrix was run.
A sampling-based controller is also available as a plug-in but is not part of the reported results; its presence indicates the interface rather than a tested capability.

\subsection{Execution, Logging, and Inspection}
\label{sec:execution}

A scheduler keeps three rates distinct: physics at 1 ms, control at 20 ms, and event and state logging at 10 ms.
Commands are held between control updates, states at a control instant are sampled before the new command is applied, and the terminal state is retained, which makes an event time comparable across a command, a state, and a task decision.

Interactive display, batch execution, and offline analysis share the state and control components but carry different responsibilities: the first renders a session, the second schedules many runs with workers using identical numerical settings, and the third reads stored records without re-simulating.
A record holds the scenario, backend, and configuration identifiers, the commanded controls, the logged states, and the terminal event, so an analyst can locate one outcome, inspect the first violation, and compare the response preceding it.

%% file: sections/04_protocol.tex
\section{Task-Oriented Evaluation}
\label{sec:protocol}

The evaluation method consumes the platform's records.
It defines what a task requests, what counts as completion, and how a failure is examined.

\subsection{Task Families and Reference Demand}
\label{sec:tasks}

\begin{figure*}[t]
\centering
\includegraphics[width=0.92\textwidth]{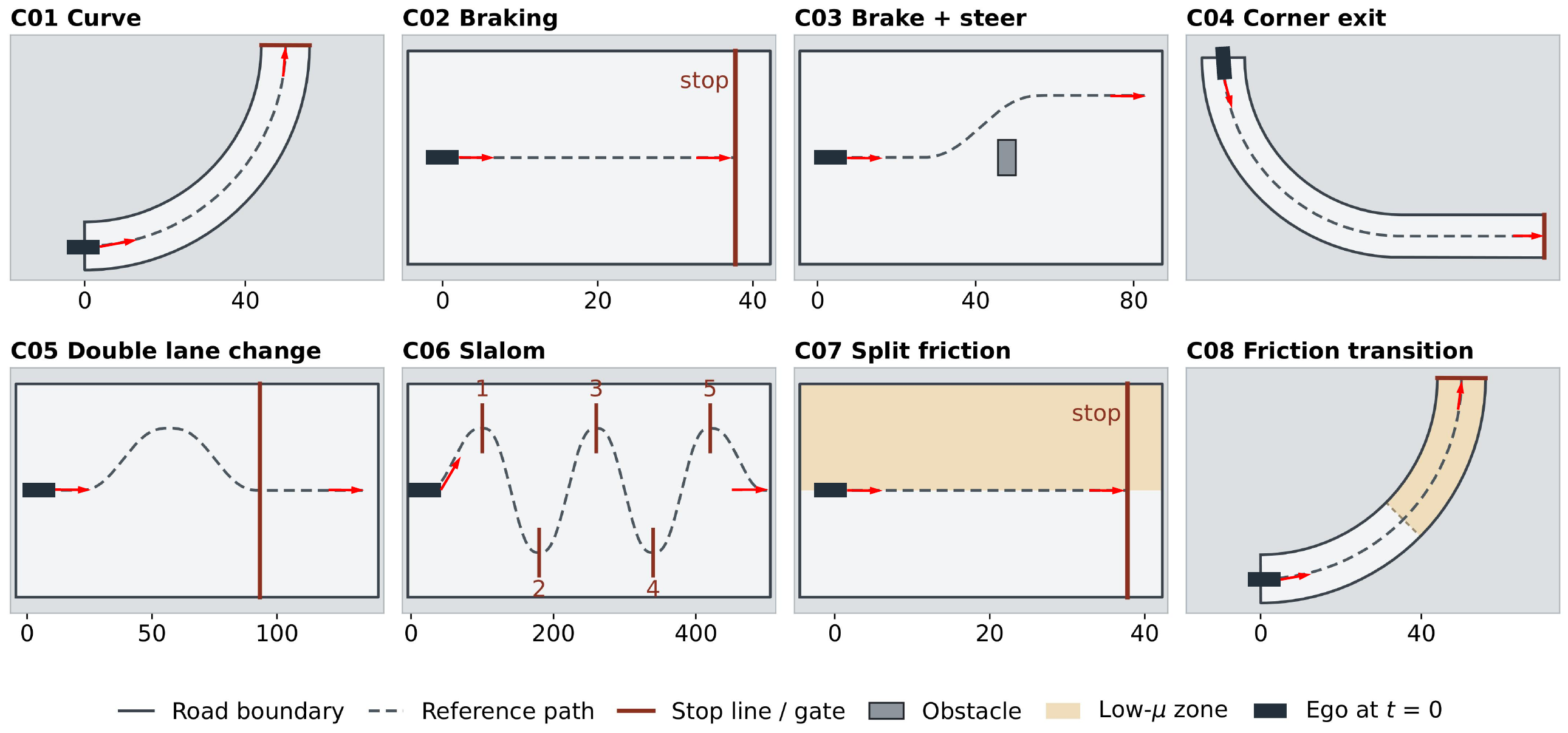}
\caption{Task-family sketches in the order C01--C08 of Table~\ref{tab:tasks}, taken from one stored definition per family (M02, L03): road boundary (gray), reference polyline (blue, dashed), task mark (orange: stop line or ordered gate), obstacle (brown), and the friction divider (thin line) for C07 and C08.
Panel scale and orientation follow each family's geometry, so a level is comparable within a family but not across families; no ego vehicle is drawn.
}
\label{fig:families}
\end{figure*}

Each task instance specifies the road, obstacles, friction field, initial state, geometric reference, speed schedule, completion goal, and deadline.
The eight families cover cornering, braking, combined braking and steering, corner exit, lane change with recovery, slalom, split-friction braking, and a friction transition (Table~\ref{tab:tasks}, Fig.~\ref{fig:families}).
For a controller-independent reference with speed $v_{\mathrm{ref}}$ and curvature $\kappa_{\mathrm{ref}}$, let $a_{x,\mathrm{ref}}=\dot v_{\mathrm{ref}}$ and $a_{y,\mathrm{ref}}=v_{\mathrm{ref}}^2\kappa_{\mathrm{ref}}$.
On uniform friction $\mu>0$ the demand coordinates are
\begin{equation}
\label{eq:reference-demand}
\begin{aligned}
D_x^{\mathrm{req}}&=\max_t |a_{x,\mathrm{ref}}(t)|/(\mu g),\quad
D_y^{\mathrm{req}}=\max_t |a_{y,\mathrm{ref}}(t)|/(\mu g),\\
D_{xy}^{\mathrm{req}}&=\max_t \frac{\sqrt{a_{x,\mathrm{ref}}(t)^2+a_{y,\mathrm{ref}}(t)^2}}{\mu g}.
\end{aligned}
\end{equation}
The joint form combines accelerations at the same reference sample rather than independently attained peaks.
These quantities describe the prescribed maneuver; they are neither a measured vehicle capacity nor a lower bound for every strategy, and a value above one does not establish infeasibility.

For C01--C06, environments M01--M03 use uniform friction $\mu\in\{0.3,0.6,0.9\}$ and five target levels $\{0.5,0.7,0.9,1.1,1.3\}$ in the family coordinate of Table~\ref{tab:tasks}.
The generator changes speed or maneuver geometry to realize a level and the realized value is recomputed from the sampled reference, so a level is meaningful within its family rather than as a universal difficulty score.
C07 varies friction asymmetry about a fixed nominal braking demand $D_b=0.9$ as $\mu_L=\bar\mu(1-A_\mu)$ and $\mu_R=\bar\mu(1+A_\mu)$ with $\bar\mu\in\{0.3,0.6,0.9\}$ and $A_\mu\in\{0.10,0.25,0.40,0.55,0.70\}$.
C08 varies $R_\mu=\mu_{\mathrm{after}}/\mu_{\mathrm{before}}\in\{0.90,0.75,0.60,0.45,0.30\}$ with entry friction $\{0.9,0.7,0.5\}$.
Friction is evaluated at individual wheel positions, so left and right contacts can meet different values.

\input{tables/table1_tasks}

\subsection{Task Outcomes and Sampled Passing Sets}
\label{sec:protocol-envelope}

For a valid run of chain $\pi$, model $b$, and task $\xi$, $Y(\pi,b,\xi)=\mathrm{Pass}$ requires the goal by its deadline and compliance with road, obstacle, and task constraints through completion.
C01, C03, and C08 require a forward crossing of the oriented exit gate, C04 additionally an exit speed, C02 and C07 a stop before the line below a speed threshold, and C05 and C06 ordered gates with a sustained recovery hold where specified; goal attainment alone is insufficient.

The evaluated footprint is a yaw-rotated 4.5 m by 1.8 m rectangle.
Straight-road tasks use the declared road polygon; curved ribbons check the center and four corners against the reference-polyline half-width, with endpoint tangents extended by half the vehicle length.
Collisions use vehicle and obstacle polygons with a separating-axis test.
Checks occur at event samples and do not implement continuous swept-volume detection.

Numerical or interface faults are labeled \emph{Invalid} and missing records are listed separately; neither enters the physical-failure count.
For a finite task set $\Xi$ at fixed family, environment, and configuration, the empirical passing set is
\begin{equation}
\label{eq:empirical-capability}
\widehat{\mathcal C}_{\pi,b}=\{\xi\in\Xi:Y(\pi,b,\xi)=\mathrm{Pass}\}.
\end{equation}
Outcome maps display tested cells.
Adjacent changes identify intervals for additional sampling, and nonmonotonic results are retained, because the set is conditional on the model, configuration, and finite sample rather than a continuous stability region.

\subsection{Event-Aligned Diagnosis}
\label{sec:diagnosis}

Failures are classified by the first task-violating event as collision, road departure, stop-line violation, recovery timeout, or incomplete task, with that order as a fixed tie-breaker for simultaneous sampled events; all events are retained.
Online termination and offline evaluation apply the same rules, including the terminal sample, so re-evaluation tests record consistency rather than independent model validity.

Records retain target and actual longitudinal speed, commanded steering and pedals, yaw rate $r$, body sideslip $\beta=\operatorname{atan2}(v_y,v_x)$, and wheel-force utilization
\begin{equation}
U_i=\frac{\sqrt{F_{x,i}^2+F_{y,i}^2}}{\mu_i F_{z,i}}.
\end{equation}
Sideslip summaries use samples with speed at least 1 m/s; tire samples require finite channels, $F_z>50$ N, and $\mu>0$, and missing initial forces are not replaced by zero.
For a pass/fail pair, peak summaries share the interval $0\le t<t_f$ and exclude the first-failure sample, and steering logs contain commands rather than measured actuator angles.
These conventions make a comparison reproducible; they do not turn temporal association into a causal mechanism.

%% file: tables/table1_tasks.tex
\begin{table}[t]
\caption{Task families and primary sweep variables.
Success also requires no collision or off-road event and completion within the task limit.}
\label{tab:tasks}
\centering
\small
\setlength{\tabcolsep}{2.5pt}
\renewcommand{\arraystretch}{1.08}
\begin{tabular}{p{0.34\columnwidth}cp{0.39\columnwidth}}
\toprule
Task family & Sweep & Success condition \\
\midrule
C01 Curve & $D_y$ & Reach curve exit \\
C02 Braking & $D_b\;(=D_x)$ & Stop before line \\
C03 Brake + steer & $D_{xy}$ & Avoid obstacle; reach exit \\
C04 Corner exit & $D_{xy}$ & Reach exit at required speed \\
C05 Double lane change & $D_y$ & Pass gates; recover \\
C06 Slalom & $D_y$ & Pass ordered gates \\
C07 Split friction & $A_\mu$ & Stop before line \\
C08 Friction transition & $R_\mu$ & Reach curve exit \\
\bottomrule
\end{tabular}
\end{table}

%% file: sections/05_experiments.tex
\section{Platform Application and Experimental Studies}
\label{sec:experiments}

\subsection{Experimental Setup}
\label{sec:exp-setup}

E1 comprises eight families, three environments, five levels, and three chains: 120 nominal tasks and 360 runs, with every chain receiving the same state, road, reference, obstacle, and known-friction information.
Pure Pursuit follows the reference with a 12 m lookahead and Stanley uses a lateral-error gain of 2.5, both as implementations of familiar tracking laws~\cite{coulter1992purepursuit,thrun2006stanley} rather than reproductions of complete systems.
The lattice chain selects a local lateral path around the road spine and tracks it with Pure Pursuit at its defaults.
A common longitudinal PID overrides their pedal outputs, reading the reference speed at projected progress, with $(K_p,K_i,K_d)=(0.25,0.05,0)$, throttle feedforward 0.15, and integral bound 2.0; stopping tasks request zero speed immediately.

E2 inserts target demands 0.75, 0.80, and 0.85 for every chain in C01 and C03 at M02, where E1 shows a transition between 0.7 and 0.9, adding 18 runs; C01 changes speed along the existing curve, C03 regenerates maneuver length, braking reference, and obstacle position, and realized demand is recomputed in both.
The obstacle at the C03 center level 0.80 is a $4.5\times2.0$ m static rectangle at $(47.93,0.0)$ m on a braking-and-steering reference in M02, the configuration whose stored record Section~\ref{sec:case} re-examines.
Before E2/E3 outcomes were observed, E3 fixed target demands 0.5, 0.8, and 1.3 and initial offsets $\{-0.2,-0.1,0.1,0.2\}$ m, translating the starting pose along the initial reference normal with heading, velocity, road, reference, and task unchanged.
This adds 72 runs, with zero-offset records reused from E1 or E2.
Table~\ref{tab:evidence} separates the capabilities these runs cover from those they do not.

\input{tables/table2_evidence}

\begin{figure*}[t]
\centering
\includegraphics[width=0.82\textwidth]{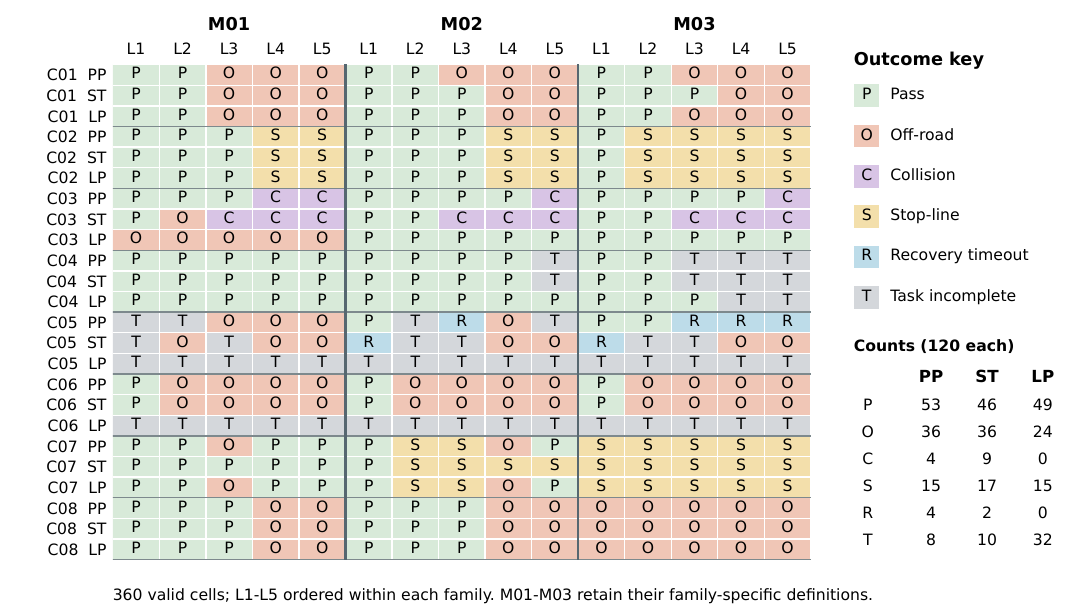}
\caption{E1 outcome map: all 360 measured cells and their first-failure categories.
Each family contains 15 task conditions per chain, and environment and level meanings are family-specific.
}
\label{fig:main}
\end{figure*}

\subsection{Batch Evaluation Across Task Families}
\label{sec:main-results}

E1 produced 148 passes and 212 task failures; PP, ST, and LP complete 53, 46, and 49 of 120 tasks, that is 44.2\%, 38.3\%, and 40.8\% of the tested cells.
These are finite-grid proportions, not estimated success probabilities.
Three observations follow from Fig.~\ref{fig:main}.

First, method differences depend on the family: ST completes more C01 tasks than PP (8 versus 6), whereas PP completes more C03 tasks than ST (11 versus 5), and LP completes 13 C04 tasks but no C05 or C06 task, so a single ranking would hide the exchange.
Second, equal totals can correspond to different cells: in C07 all three chains complete six tasks, yet for PP and LP in M01, L02 passes, L03 fails, and L04 passes again, and reporting only the largest passing level would conceal the pattern.
Third, failure categories separate chains that totals make similar: across E1 they are 96 road departures, 47 stop-line violations, 50 incomplete tasks, 13 collisions, and 6 recovery timeouts, whereas C02 shows identical cells for all chains, consistent with the shared stopping controller.
These statements concern the implemented chains and task rules, not the best achievable performance of the methods.

\subsection{Local Refinement and Initial-State Sensitivity}
\label{sec:local-results}

E2 yields 15 passes and 3 task failures.
The C01 PP transition narrows to adjacent realized demands 0.8500 and 0.9000, and the C03 ST transition to 0.7000 and 0.7529; at the inserted C03 level 0.75 ST departs the road, while at 0.80 and 0.85 its first failure is a collision.
The other chains show no transition inside the selected intervals, so refinement increases the resolution of two sampled transitions without locating a continuous boundary.

E3 yields 48 passes and 24 failures, and all 72 offset runs retain the decision of the corresponding zero-offset run.
At low demand all three chains pass both families; at target 0.8 all pass C01 while ST fails C03; at target 1.3 all fail C01 while LP passes C03.
The middle level lies inside the original coarse interval and was not chosen as a subsequently refined boundary; because the offsets are deterministic and few, the unchanged decisions do not establish robustness at the refined endpoints or a random perturbation success rate.

\begin{figure*}[t]
\centering
\includegraphics[width=0.82\textwidth]{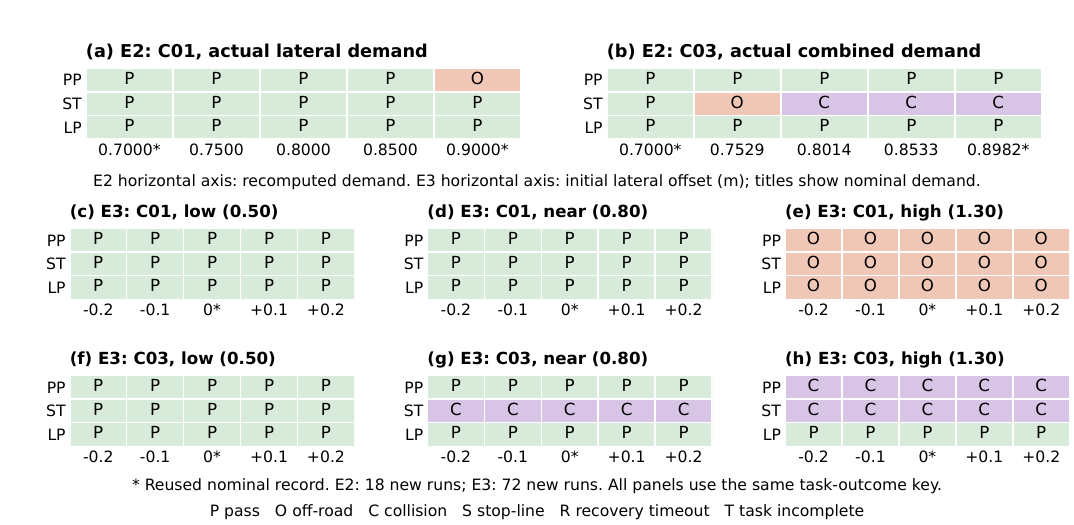}
\caption{Local refinement (E2) and deterministic initial-position sensitivity (E3), both at M02.
Ticks show realized reference demand and asterisks mark reused endpoint or zero-offset records.
The two views display 30 and 90 cells drawn from 114 unique records: 90 new runs plus 24 reused E1 runs, with six E2 center records in both views.
}
\label{fig:local}
\end{figure*}

\begin{figure*}[t]
\centering
\includegraphics[width=0.82\textwidth]{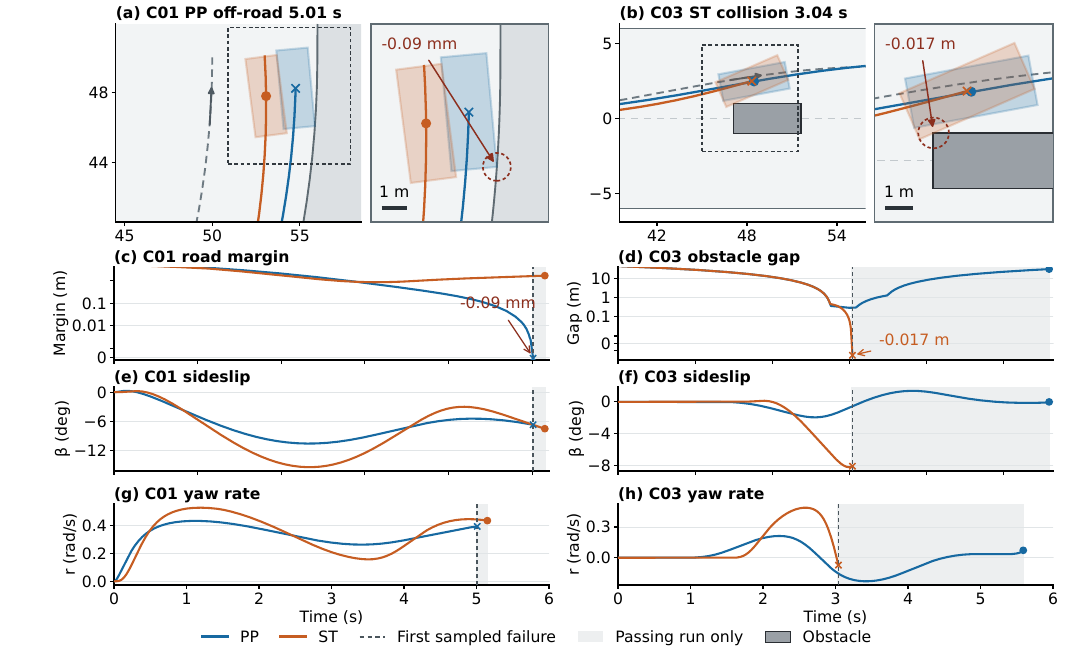}
\caption{Event-aligned records for the two cases.
(a), (b) Top view at the first-failure sample; dashed boxes mark the insets and the arrow gives the driving direction.
Gray shading is outside the 6 m corridor; the gray rectangle in (b) is the obstacle.
Dashed circles mark the annotated event; scale bars: 1 m.
(c), (d) Geometry margin on symmetric-log axes.
(e)--(h) Signed body sideslip and yaw rate; vertical lines mark the first sampled failure (5.01 s and 3.04 s), gray spans contain only the passing run, and traces end at their own terminal samples.
}
\label{fig:diagnostics}
\end{figure*}

\subsection{Event-Aligned Case Analysis}
\label{sec:case}

Two same-scenario PP/ST pairs are examined from the stored records: the first M02 C01 level with differing outcomes in E1 (L03), and the fixed center of the C03 refinement at target 0.8.
Both were selected using known outcomes, so they illustrate rather than sample the failure population.

In C01, PP first violates the sampled road constraint at 5.01 s while ST completes the exit at 5.15 s.
Defining the road margin as half-width minus the largest distance of the center and four corners to the extended reference polyline, PP changes from 0.000495 m at 5.00 s to $-0.000091$ m at 5.01 s, whereas ST retains 1.570942 m.
The 0.09 mm crossing is a discrete geometric-threshold event, not a large physical departure.
Over the common interval $[0,5.01)$ s, ST shows the larger maximum absolute sideslip (15.368$^\circ$ versus 10.531$^\circ$) and yaw rate (0.5251 versus 0.4310 rad/s) and still passes, so diagnostic amplitude alone does not reproduce the task label.

In C03, ST first collides at 3.04 s while PP completes at 5.59 s.
At 3.03 s the lateral center positions are nearly equal (2.4557 versus 2.4453 m) but the headings differ (0.4128 versus 0.1849 rad), and the normalized separating-axis gaps are 0.002609 and 0.296052 m.
At 3.04 s the ST gap becomes $-0.016997$ m, consistent with footprint overlap, while PP retains positive separation; the gap is not a Euclidean clearance.
ST's maximum wheel utilization before the event is 0.9233, so the collision label does not require utilization to reach one.
\subsection{Scope and Limitations}
\label{sec:limits}

The passing sets are conditional on one uncalibrated bicycle model, finite reference families, declared task thresholds, and fixed controller defaults.
No result here validates the model against an external solver or a physical vehicle, and the same implementation decides online and offline outcomes, so agreement tests consistency rather than independent correctness.
Sampled corner and polygon checks do not establish continuous collision avoidance, and small margins are sensitive to discretization and threshold choice.
Task failures can also reflect the baseline's task-following behavior: LP's incomplete C05 and C06 results do not show that those maneuvers are physically impossible.
The limited offsets do not test geometric generalization, stochastic robustness, sensor uncertainty, or unrestricted planning.

%% file: tables/table2_evidence.tex
\begin{table}[t]
\caption{Platform capabilities and the evidence this paper provides.
The last column states how far the reported 450 runs support each row.}
\label{tab:evidence}
\centering
\footnotesize
\setlength{\tabcolsep}{3pt}
\renewcommand{\arraystretch}{1.08}
\begin{tabular}{p{0.29\columnwidth}p{0.44\columnwidth}p{0.17\columnwidth}}
\toprule
Capability & Interface and evidence & Coverage \\
\midrule
Scenario composition & road, reference, obstacle, and friction objects; eight families over three environments & quantitative \\
Algorithm attachment & shared observation and control plug-in; three chains over 120 tasks & quantitative \\
Vehicle adaptation & adapter with selectable backend; bicycle dynamic model for every reported run & one backend \\
Interactive operation & Edit, Live, and Replay over shared components; one task illustrated (Fig.~\ref{fig:interactive}) & qualitative \\
Batch execution and analysis & scheduler and synchronized records; 450 runs with consistent online and offline outcomes & quantitative \\
Interactive agents & pluggable agent policies; no run in the reported matrix & not tested \\
\bottomrule
\end{tabular}
\end{table}

%% file: sections/05_conclusion.tex
\section{Conclusion}
DynLimit organizes dynamics-critical experiments as one process: scenarios and references are declared as objects, algorithms and vehicles attach through shared interfaces, execution keeps control, state, and event records on a single timeline, and a task-oriented method turns those records into passing sets and event-aligned diagnoses.
On this platform, 450 runs across eight families show that method differences depend on the task family, that equal aggregate counts can hide different passing cells, and that local refinement and prescribed offsets change the resolution of a result without establishing a continuous boundary.
The case comparisons confirm that geometry and completion rules carry information that response peaks alone do not.
Future work will first strengthen perception testing: sensor models, detection latency, and estimator error enter as further declared objects, so that degraded and occluded sensing is swept by the same task matrix that now sweeps demand and initial offset.
Beyond that, calibrated vehicle transfer, broader scenario coverage, and continuous-boundary estimation remain open, and the platform is the instrument we expect to use for them.